\def\confName{CVPR}
\def\confYear{2027}

\def\paperTitle{Fundus-R1: Training a Fundus-Reading MLLM on Public Data}

\def\authorBlock{
    Yuchuan Deng$^{1}$ \quad
    Qijie Wei$^{1}$ \quad
    Kaiheng Qian$^{1}$ \quad
    Jiazhen Liu$^{2}$ \quad
    Bangxiang Lan$^{1}$\\
    Qingwei Shen$^{1}$ \quad
    Jianfeng Dong$^{3}$ \quad
    Weihong Yu$^{4}$ \quad
    Xirong Li$^{1}$%
    \thanks{Corresponding author: \texttt{xirong@ruc.edu.cn}.}\\[3pt]
    $^{1}$Renmin University of China \quad
    $^{2}$The Hong Kong University of Science and Technology\\
    $^{3}$Zhejiang Gongshang University \quad
    $^{4}$Peking Union Medical College Hospital
}

\newif\ifreview 
\newif\ifarxiv \newcommand{\arxiv}{\arxivtrue}
\newif\ifcamera 
\newif\ifrebuttal 

\arxiv
\pdfoutput=1
\documentclass[10pt,twocolumn,letterpaper]{article}
\ifreview \usepackage[review]{cvpr} \fi
\ifarxiv \usepackage[pagenumbers]{cvpr} \fi
\ifrebuttal \usepackage[rebuttal]{cvpr} \fi
\ifcamera \usepackage{cvpr} \fi

\usepackage{graphicx}	
\usepackage{amsmath}	
\usepackage{amssymb}	
\usepackage{booktabs}
\usepackage{times}
\usepackage{microtype}
\usepackage{epsfig}
\usepackage{caption}
\usepackage{float}
\usepackage{placeins}
\usepackage{color, colortbl}
\usepackage{stfloats}
\usepackage{enumitem}
\usepackage{tabularx}
\usepackage{xstring}
\usepackage{multirow}
\usepackage{xspace}
\usepackage{url}
\usepackage{subcaption}
\usepackage{xcolor}
\usepackage[hang,flushmargin]{footmisc}

\ifcamera \usepackage[accsupp]{axessibility} \fi

\ifarxiv  \fi

\newcommand{\R}[1]{{%
    \textbf{%
        \ifstrequal{#1}{1}{\textcolor{red}{R#1}}{%
        \ifstrequal{#1}{2}{\textcolor{blue}{R#1}}{%
        \ifstrequal{#1}{3}{\textcolor{magenta}{R#1}}{%
        \ifstrequal{#1}{4}{\textcolor{teal}{R#1}}{%
                           \textcolor{cyan}{R#1}%
        }}}}%
    }%
}}
\usepackage{xr-hyper}

\makeatletter
\newcommand*{\addFileDependency}[1]{
  \typeout{(#1)}
  \@addtofilelist{#1}
  \IfFileExists{#1}{}{\typeout{No file #1.}}
}

\makeatother
\newcommand*{\myexternaldocument}[1]{
    \externaldocument{#1}
    \addFileDependency{#1.tex}
    \addFileDependency{#1.aux}
}

\definecolor{cvprblue}{rgb}{0.21,0.49,0.74}
\definecolor{linkred}{rgb}{1.0,0,0}
\usepackage[pagebackref,breaklinks,colorlinks,
    linkcolor=linkred, % 图表引用为红色
    citecolor=cvprblue % 文献引用为蓝色
]{hyperref}
\usepackage[capitalize]{cleveref}
\crefname{section}{Sec.}{Secs.}
\crefname{table}{Table}{Tables}
\crefname{figure}{Fig.}{Figs.}

\ifarxiv \crefname{appendix}{App.}{Apps.}
\else \crefname{appendix}{Suppl.}{Suppls.} \fi

\unless\ifarxiv \myexternaldocument{_supplementary} \fi

\usepackage{xspace}
\usepackage{amsmath,amsfonts}
\usepackage{graphicx}
\usepackage{booktabs}
\usepackage{microtype}
\usepackage{nicefrac}
\usepackage{ragged2e}
\usepackage{placeins}
\usepackage{wrapfig}

\usepackage[table]{xcolor}
\usepackage{makecell}
\usepackage{multirow}
\usepackage{subcaption}
\usepackage{siunitx}
\usepackage{tabularx}
\usepackage{enumitem}
\usepackage{bbm}
\usepackage{pifont}

\newcommand{\cmark}{\ding{51}} % ✓
\newcommand{\xmark}{\ding{55}} % ✗
\newcommand{\model}{\texttt{Fundus-R1}\xspace}
\newcolumntype{Y}{>{\raggedright\arraybackslash}X}
\newcolumntype{C}[1]{>{\centering\arraybackslash}m{#1}}

\definecolor{famspec}{RGB}{78,121,167}   % U-Net / specialist  (blue)
\definecolor{famsam}{RGB}{242,142,43}    % SAM-based           (orange)
\definecolor{famclip}{RGB}{89,161,79}    % CLIP-based          (green)
\definecolor{fammllm}{RGB}{176,122,161}  % MLLM-based          (purple)
\definecolor{famsup}{RGB}{225,87,89}     % support-conditioned (red)
\crefname{section}{Sec.}{Secs.}
\Crefname{section}{Sec.}{Secs.}

\crefname{table}{Tab.}{Tabs.}
\Crefname{table}{Tab.}{Tabs.}

\crefname{figure}{Fig.}{Figs.}
\Crefname{figure}{Fig.}{Figs.}

\definecolor{cvprblue}{RGB}{33,113,181}
\definecolor{cvprorange}{RGB}{230,85,13}
\definecolor{cvprgreen}{RGB}{49,163,84}
\definecolor{cvprgray}{RGB}{110,110,110}
\definecolor{lightgray}{RGB}{230,230,230}

\begin{document}

\title{\paperTitle}
\author{\authorBlock}
\maketitle

\begin{abstract}

Fundus imaging such as CFP, OCT and SLO is crucial for the early detection of retinal anomalies and diseases. Fundus image understanding, due to its knowledge-intensive nature, poses a challenging vision-language task. An emerging approach to addressing the task is to post-train a generic multimodal large language model (MLLM), either by supervised finetuning (SFT) or by reinforcement learning with verifiable rewards (RLVR), on a considerable amount of in-house samples paired with high-quality clinical reports. However, these valuable samples are not publicly accessible, which not only hinders reproducibility but also practically limits research to few players. In this paper, we make a novel attempt to train a \emph{reasoning}-enhanced fundus-reading MLLM, termed \model, using exclusively \textbf{public} datasets, wherein over 94\% of the data are annotated with image-level labels only. Our technical contributions are two-fold. First, we propose a RAG-based method for composing image-specific, knowledge-aware reasoning traces. Such auto-generated traces link visual findings identified by a generic MLLM to the image labels in terms of ophthalmic knowledge. Second, we enhance RLVR with a process reward that encourages self-consistency of the generated reasoning trace in each rollout. Extensive experiments on three fundus-reading benchmarks, i.e., FunBench, Omni-Fundus and GMAI-Fundus, show that \model clearly outperforms multiple baselines, including its generic counterpart (Qwen2.5-VL) and a stronger edition post-trained without using the generated traces. This work paves the way for training powerful fundus-reading MLLMs with publicly available data.
\end{abstract}

\section{Introduction} 
\label{sec:intro}

The retina, our body's sole visual sensor, is primarily examined via fundus imaging. Color fundus photography (CFP) offers an en face view of the retina, while optical coherence tomography (OCT) provides cross-sectional visualizations \cite{mm21-mm-mil}. The ability of retinal experts to read these images -- identifying lesions, discerning their type, location, and quantity, and subsequently diagnosing retinal diseases -- is a critical skill cultivated through years of rigorous practice. This paper aims to imbue a multimodal large language model (MLLM) with this skill. The resultant MLLM, which we term \model, is obtained with novel \emph{reasoning}-enhanced training on existing public datasets currently \emph{lacking} reasoning-trace annotations for supervised learning.

Fundus image understanding is inherently knowledge-intensive, making it a challenging vision-language task.
Unlike common image recognition, fundus reading requires the model not only to identify (subtle) \emph{\textbf{visual findings}} (\textbf{VF}) in a given fundus image, but also to leverage highly specific domain knowledge to decode these findings into diagnostic output. 
% Unlike common image recognition, fundus reading requires the model not only to identify (subtle) visual findings in a given fundus image, but also to leverage highly specific domain knowledge to decode these findings into diagnostic output. 
Accordingly, recent efforts on this task have increasingly focused on post-training a generic MLLM into a powerful fundus-image reader \cite{li2024visionunite,fundusexpert,li2025eyecaregpt,OphthaReason}.

\begin{table}[t]
\caption{\textbf{\model \emph{versus} existing fundus-reading MLLMs}. The star symbol (*) indicates models that are publicly accessible and thus evaluated in our experiments. Our model is exclusively trained on public data, with over 94\% of the training samples having image-level labels only. }
\label{tab:data_reward_compare2}
\setlength{\tabcolsep}{2.5pt}
\renewcommand{\arraystretch}{1.0}
\resizebox{\linewidth}{!}{%
\begin{tabular}{@{}l c c r@{}}
\toprule
\textbf{Model} &
\textbf{\makecell[c]{Learned from \\clinical reports?}} &
\textbf{\makecell[c]{Reasoning\\trace}} &
\textbf{RLVR Reward} \\
\midrule
DeepDR-LLM~\cite{DeepDR-LLM}
& \cmark (private)
& \xmark
& -- \\
VisionFM~\cite{qiu2024development}
& \cmark (private)
& \xmark
& -- \\
VisionUnite~\cite{li2024visionunite}
& \cmark (private)
& \cmark
& -- \\
RetinalGPT~\cite{zhu2025retinalgpt}
& \cmark (private)
& \cmark
& -- \\
EyeCareGPT~\cite{li2025eyecaregpt}
& \cmark (private)
& \cmark
& -- \\
FundusExpert*~\cite{fundusexpert}
& \cmark (private)
& \cmark
& -- \\
OphthaReason*~\cite{OphthaReason}
& \cmark (public)
& \cmark
& Answer + Format \\
%\midrule
\rowcolor{blue!20}
\textbf{\model{}}
& \xmark
& \cmark
& Answer + Format + \textbf{Process} \\
\bottomrule
\end{tabular}%
}
\centering

\end{table}

Current fundus-reading MLLMs mostly rely on richer supervision constructed from a mixture of public and private resources, such as in-house samples paired with high-quality clinical reports or private retinal images, see Tab.~\ref{tab:data_reward_compare2}. In particular,  
% {Even when public data are partially used, 
\textbf{the most informative supervision in these methods largely comes from the private resources}. 
While such resources have substantially advanced the field, % ophthalmic multimodal modeling, 
they are unfortunately not publicly accessible. 
Such a limitation not only weakens research reproducibility but, more importantly, restricts further progress to a small number of groups that have privileged access to proprietary data.
Some initial efforts have been made to gather clinical reports from public resources. 
% For instance, OphthaReason \cite{OphthaReason} attempts to extract such reports from PubMed Central (PMC)\footnote{\url{https://pmc.ncbi.nlm.nih.gov/}}. 
For instance, OphthaReason \cite{OphthaReason} attempts to extract such reports from PubMed Central (PMC) \cite{nlm_pmc}. 
However, as these reports are primarily released for educational purposes, their quantity is inherently limited.

%A fundamental challenge  that publicly available fundus datasets are themselves highly weakly supervised. As shown in \cref{tab:datasets}, our collected public training pool contains 171,420 samples in total, yet over 93\% of them provide only image-level labels, while only a small minority include pixel-level lesion annotations. Such supervision may tell the model what the answer is, but provides much less guidance on how \emph{visual evidence} should be organized into a diagnostic reasoning trace. This raises a key question for our work: can a reasoning-enhanced fundus-reading MLLM \model be trained using exclusively public datasets, even when the vast majority of available samples provide only image-level labels?

When relying on  training with public fundus-image datasets, a major challenge is that most of these datasets provide only image-level labels, see Tab.~\ref{tab:datasets}. 
When learning from publicly accessible datasets, a major challenge is that most of the datasets provide only image-level labels, see the supplement.
Such holistic supervision may inform the model of the correct answer, but provides little guidance on how VFs from fundus images should be organized into a diagnostic reasoning trace. This raises a key  question: \textbf{can a reasoning-enhanced fundus-reading MLLM be trained using exclusively public data?}

\begin{table}[!htbp]
\caption{\textbf{Training data sources for \model}. 
To avoid data leakage, images included in our test sets, \ie FunBench~\cite{wei2025funbench}, Omni-Fundus \cite{hu2024omnimedvqa} and GMAI-Fundus \cite{yegmai}, are excluded. Over 94\% of our training samples are with image-level labels only.}
\label{tab:datasets}
\setlength{\tabcolsep}{4pt}

\centering
\resizebox{\linewidth}{!}{%
\begin{tabular}{@{}l r r l@{}}
\toprule
\textbf{Dataset}  & \textbf{\#Samples}  & \textbf{Pixel labeled} & \textbf{Primary tasks} \\
\midrule

\multicolumn{4}{@{}l}{\emph{Image modality: CFP}} \\
FGADR-143-9~\cite{kheradfallah2022annotation} & 143     & 143   & DR-specific lesion segmentation \\
IDRiD~\cite{idrid}                      & 413      &54     & DR grading \\
GRAPE~\cite{huang2023grape}             & 631     & 0     & Glaucoma progression prediction \\
JSIEC~\cite{jsiec}                      & 800     & 0     & 35-class disease classification \\
Retinal-Lesions~\cite{retinal-lesions}  & 1,264   & 1,264 & DR grading \\
RFIMiD~\cite{rfmid}                     & 2,560   & 0     & 46-label disease classification \\
OIA-ODIR~\cite{oia-odir}                & 7,342   & 0     & 8-label disease classification \\
DDR~\cite{ddr}                          & 8,763    &531   & DR grading \\

\multicolumn{4}{@{}l}{\emph{Image modality: OCT}} \\
OCTID~\cite{octid}                      & 458     & 0     & 5-class disease classification \\
OCTDL~\cite{octdl}                      & 1,651   & 0     & 7-class classification and grading \\
OIMHS~\cite{ye2023oimhs}                & 3,859   & 3859     & 4-class lesion segmentation \\
RETOUCH~\cite{retouch}                  & 5,697   & 2742     & Fluid segmentation \\
%全阴也认为是mask
NEH~\cite{neh}                          & 13,642  & 0     & 3-class disease classification \\
UCSD~\cite{UCSD}                        & 107,312 & 0     & 4-class disease classification \\ 

\multicolumn{4}{@{}l}{\emph{Image modality: SLO}} \\
TOP \cite{top}                                     & 10,433  & 0     & 9-label disease classification \\ 

\multicolumn{4}{@{}l}{\emph{Image modalities: CFP + SLO}} \\
DeepDRiD~\cite{deepdrid}                & 1,800   & 0     & DR grading \\ [3pt]

\multicolumn{4}{@{}l}{\emph{Image modalities: CFP + OCT}} \\
MMC-AMD~\cite{mmc-amd2}                 & 2170     & 0     & 4-class AMD classification \\
\midrule
\textbf{Total} & \textbf{ 168,938} & \textbf{ 8593} & \textbf{--} \\
\bottomrule
\end{tabular}%
}

\end{table}

%A straightforward idea is to use a generic MLLM to first construct reasoning traces from existing image-label pairs, and then use these traces to post-train a base model,  the model further~\cite{xu2024llava,yao2024mulberry}. However, this route faces a challenge: generic MLLMs lack sufficient ophthalmic domain knowledge~\cite{wei2025funbench,qin2024lmod}. As a result, if reasoning traces are synthesized directly from the image and label alone, the generated outputs often suffer from insufficient visual evidence, improper use of domain knowledge, or logical flawed reasoning chains (see \cref{tab:show-traces}), making them unreliable as intermediate supervision. 

\begin{table*}[t]
\centering
\caption{
\textbf{Reasoning traces generated for given VQA triplets}.
Free-form CoT generation tends to introduce \textcolor{red}{over-interpretation and logically inconsistent} reasoning,
while our method produces \textcolor{blue}{evidence-aligned and logically consistent} reasoning. Best viewed in color.
}
%\vspace{-1.5mm}

\begingroup
\scriptsize
\setlength{\parskip}{0pt}
\setlength{\parindent}{0pt}

\providecommand{\traceheader}{%
\noindent
\begin{minipage}[t]{0.205\textwidth}
\raggedright\bfseries Input VQA
\end{minipage}\hfill
\begin{minipage}[t]{0.385\textwidth}
\raggedright\bfseries Free-form CoT generation (Qwen3-VL-Plus)
\end{minipage}\hfill
\begin{minipage}[t]{0.385\textwidth}
\raggedright\bfseries Knowledge-aware generation (Ours)
\end{minipage}\par
\vspace{0.25em}\hrule\vspace{0.35em}
}

\providecommand{\tracecase}[5]{%
\noindent
\begin{minipage}[t]{0.205\textwidth}
    \vspace{0pt}
    \centering
    \includegraphics[height=1.55cm]{#1}\par\vspace{-0.4em}
    {\raggedright #2\par}
    \vspace{-0.4em}
    {\raggedright\bfseries #3\par}
\end{minipage}\hfill
\begin{minipage}[t]{0.385\textwidth}
    \vspace{0pt}
    \raggedright
    \linespread{0.96}\selectfont
    #4
\end{minipage}\hfill
\begin{minipage}[t]{0.385\textwidth}
    \vspace{0pt}
    \raggedright
    \linespread{0.96}\selectfont
    #5
\end{minipage}\par
\vspace{0.1em}\hrule\vspace{0.1em}
}

\hrule height 0.6pt
\vspace{0.3em}

\traceheader

\tracecase
{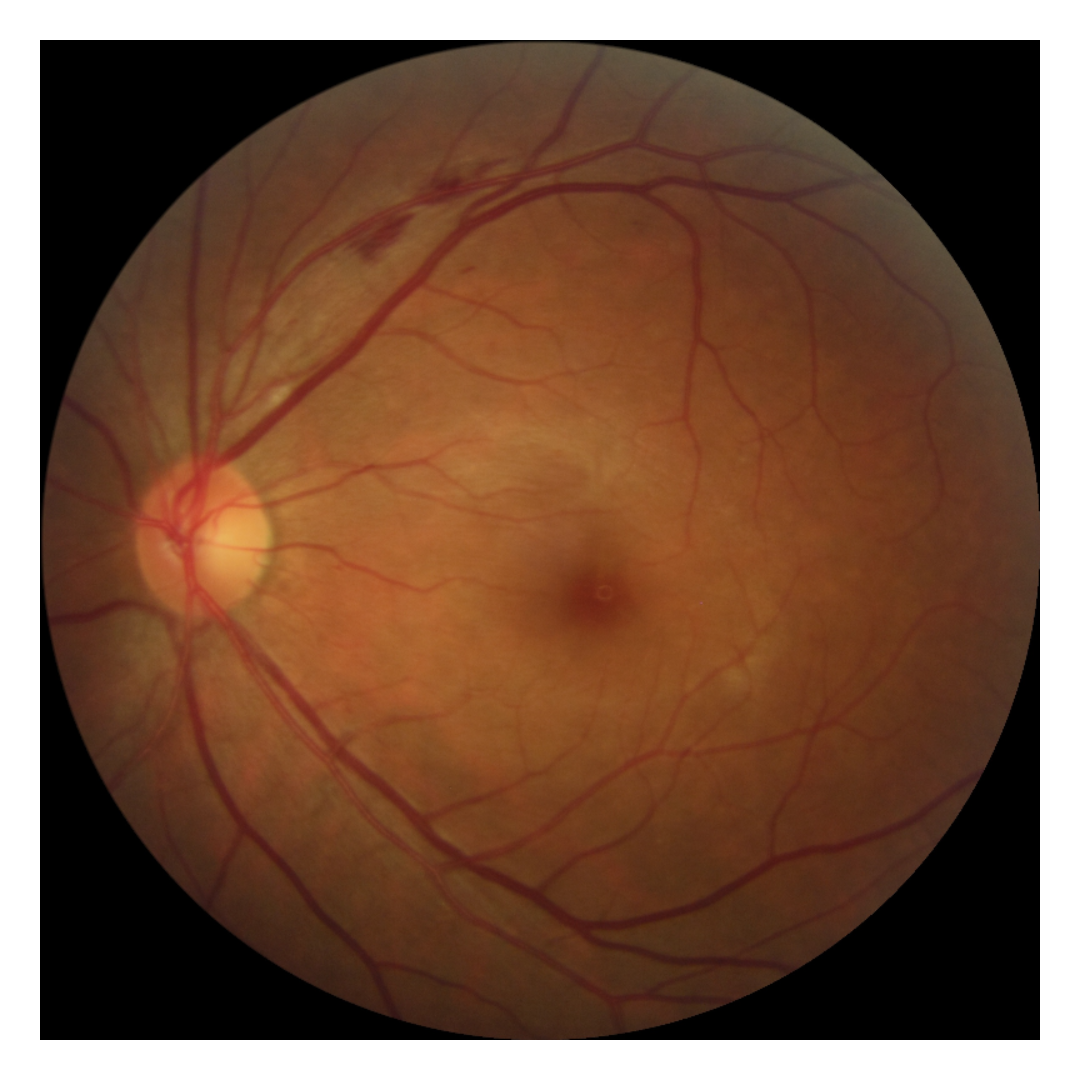}
{Which level of diabetic retinopathy is shown in the fundus image?}
{Moderate NPDR}
{%
\texttt{<think>}The fundus image shows scattered microaneurysms and dot hemorrhages around the posterior pole. Multiple hard exudates and mild cotton-wool spots are also visible. However, the mention of \textcolor{red}{\textbf{venous beading}} and \textcolor{red}{\textbf{intraretinal microvascular abnormalities}} suggests more advanced vascular damage than is actually supported by the image. Therefore, the condition is interpreted as \textcolor{blue}{\textbf{moderate NPDR}}.\texttt{</think>} \texttt{<answer>}Moderate NPDR\texttt{</answer>}%
}
{%
\texttt{<think>}The fundus image shows multiple \textcolor{blue}{\textbf{microaneurysms}} and \textcolor{blue}{\textbf{dot--blot hemorrhages}} in the posterior pole. No neovascularization, preretinal hemorrhage, or vitreous hemorrhage is observed. Compared with mild NPDR, the increased number and broader distribution of these lesions support a diagnosis of \textcolor{blue}{\textbf{moderate NPDR}}.\texttt{</think>} \texttt{<answer>}Moderate NPDR\texttt{</answer>}%
}

\tracecase
{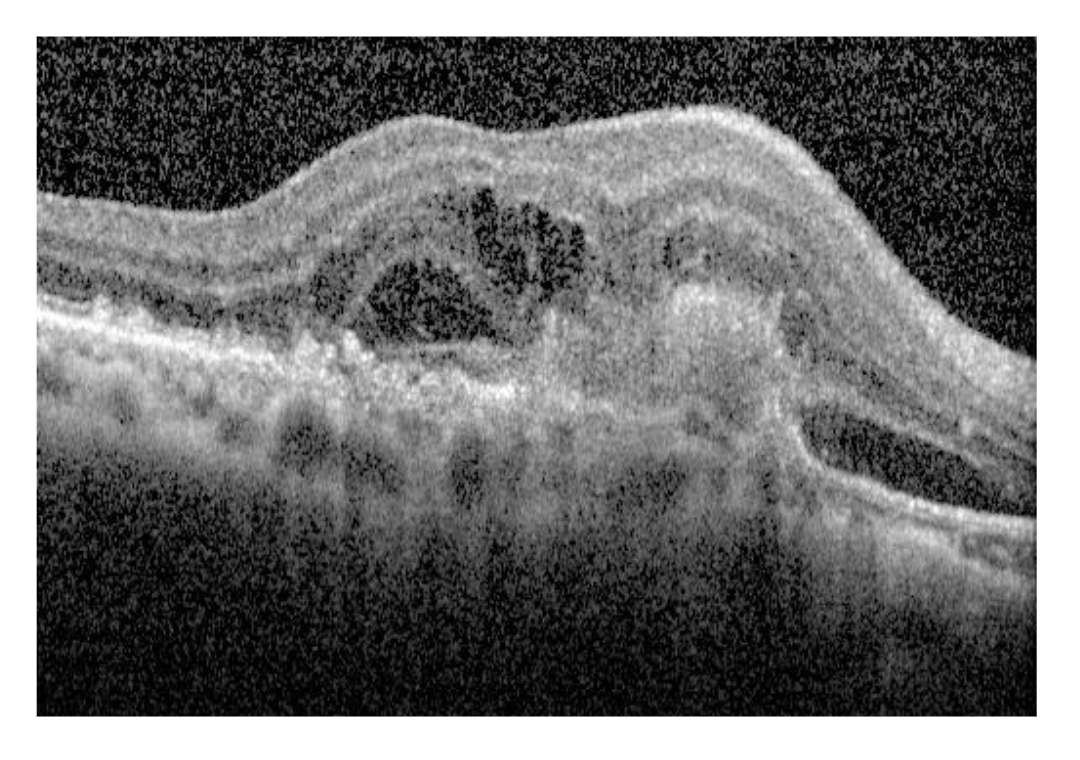}
{What type of AMD is shown here?}
{Exudative AMD}
{%
\texttt{<think>}The OCT shows elevation of the retinal pigment epithelium with subretinal hyperreflective material and mild intraretinal fluid. It also mentions outer retinal layer thinning and ellipsoid-zone disruption, but \textcolor{red}{\textbf{no clear signs of active leakage}} are claimed despite visible fluid. These findings are interpreted as \textcolor{blue}{\textbf{exudative AMD}}.\texttt{</think>} \texttt{<answer>}Exudative AMD\texttt{</answer>}%
}
{%
\texttt{<think>}The OCT demonstrates a \textcolor{blue}{\textbf{large pigment epithelial detachment (PED)}} together with \textcolor{blue}{\textbf{subretinal fluid (SRF)}} and \textcolor{blue}{\textbf{intraretinal fluid (IRF)}}. This combination is characteristic of \textcolor{blue}{\textbf{exudative AMD}} and indicates ongoing neovascular activity.\texttt{</think>} \texttt{<answer>}Exudative AMD\texttt{</answer>}%
}

\tracecase
{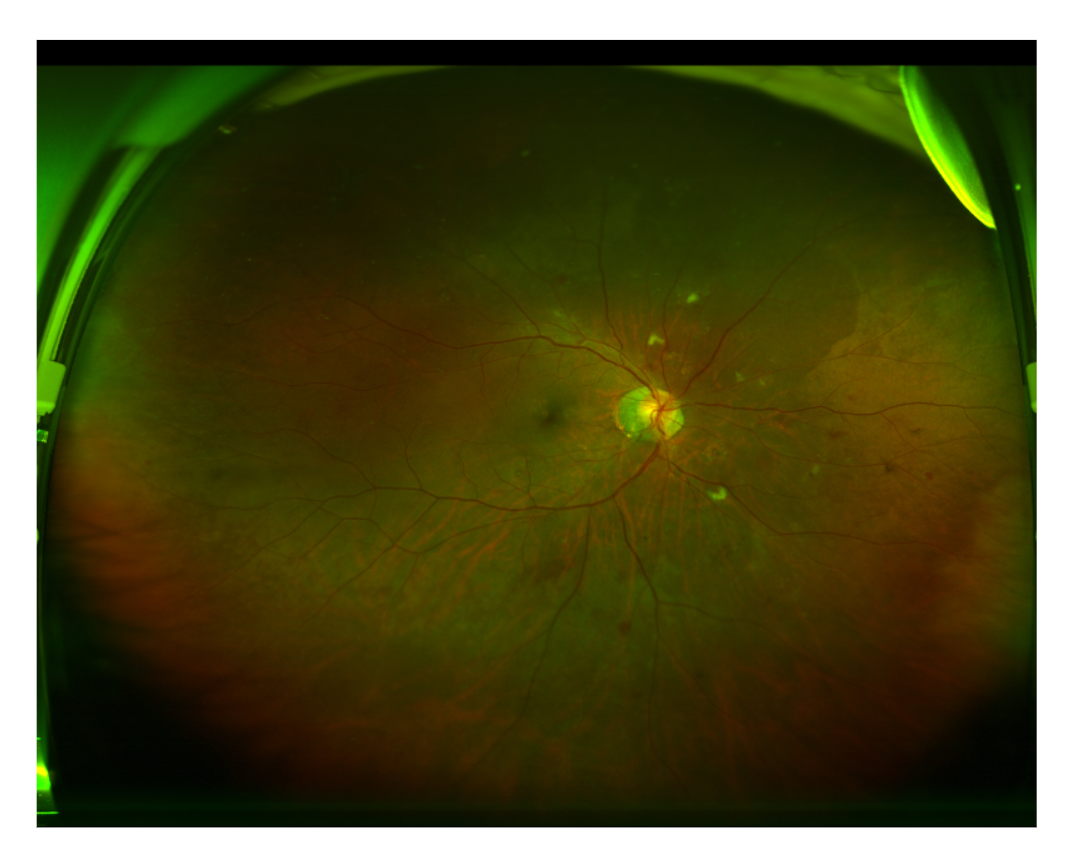}
{What fundus disease is shown?}
{Glaucoma}
{%
\texttt{<think>}The ultra-widefield fundus image shows a relatively large optic disc with an increased cup-to-disc ratio. Mild superior rim thinning is noted, and the trace further argues that \textcolor{red}{\textbf{the absence of retinal lesions}} supports glaucoma, which is not a valid diagnostic basis. The conclusion is \textcolor{blue}{\textbf{glaucoma}}.\texttt{</think>} \texttt{<answer>}Glaucoma\texttt{</answer>}%
}
{%
\texttt{<think>}The image shows an optic disc with a \textcolor{blue}{\textbf{markedly enlarged cup-to-disc ratio}}. The cup is deep and vertically elongated, with \textcolor{blue}{\textbf{superior and inferior rim thinning}} in a typical glaucomatous pattern. These findings indicate \textcolor{blue}{\textbf{structural optic nerve damage consistent with glaucoma}}.\texttt{</think>} \texttt{<answer>}Glaucoma\texttt{</answer>}%
}

\endgroup
%\vspace{-2mm}
\vspace{1em}

\label{tab:show-traces}

\end{table*}

A straightforward approach to answering this question is to prompt a generic MLLM to construct reasoning traces from existing image–label pairs, 
%\yc{A straightforward approach is to prompt a generic MLLM to construct reasoning traces from existing image–label pairs,} 
and then use these traces to post‑train a base model. 
This idea has demonstrated high potential for tasks such as item counting, geometric reasoning, and mathematical problem solving \cite{xu2024llava,yao2024mulberry}. 
However, this strategy is ineffective in the current context, as generic MLLMs lack sufficient ophthalmic domain knowledge \cite{wei2025funbench,qin2024lmod}. 
When synthesizing reasoning traces directly from the image and label, the generated outputs often suffer from insufficient visual evidence, improper use of domain knowledge, or logically flawed reasoning chains (\cref{tab:show-traces}), rendering them unreliable as intermediate supervision.

We too leverage generic models, but through a carefully designed pipeline that harnesses the ability of generic MLLMs for low-level visual recognition and that of generic LLMs for structured information extraction. 
Specifically, we first employ retrieval-augmented generation (RAG) to construct label- and modality-specific \emph{\textbf{domain knowledge}} (\textbf{DK}) from public ophthalmic references, including 
% (EyeWiki\footnote{\url{https://eyewiki.org/Main_Page}}, AAO\footnote{\url{https://www.aao.org/education}}, and PMC).
EyeWiki \cite{aao_eyewiki}, AAO Clinical Education \cite{aao_clinicaleducation}, and PMC \cite{nlm_pmc}.
From the DK, we derive a task-conditioned VF vocabulary and use a generic MLLM to extract image-specific VFs from each training image.
Based on the extracted VFs and the label-conditioned knowledge, we compose knowledge-aware and image-grounded reasoning traces. This  design not only enhances the reliability of reasoning supervision. More importantly, reasoning supervision
can now be induced from public data with mainly image-level labels, even when the underlying MLLM lacks sufficient ophthalmic knowledge. 
%We then combine the extracted VFs with the corresponding label-conditioned domain knowledge (DK) to compose knowledge-aware reasoning traces. \yc{As VF extraction serves as a modular evidence-acquisition component, the generic MLLM used here can be naturally replaced or complemented by specialized lesion detection or segmentation models to provide finer-grained visual evidence.} This design \yc{offers a practical route to inducing reasoning supervision} from public datasets dominated by image-level labels, even when the underlying MLLM has limited ophthalmic knowledge.
% From this structured knowledge, we derive a task-conditioned vocabulary of visual findings and use a generic MLLM to extract image-specific visual findings(VFs) from each training image. Based on the extracted VFs and the corresponding label-conditioned DK, we then compose knowledge-aware reasoning traces.
% This design not only enhances the reliability of reasoning supervision. More importantly, reasoning supervision can now be induced from public data with mainly image-level labels, even when the underlying MLLM lacks sufficient ophthalmic knowledge. 
To effectively leverage the reasoning-enriched training data for post-training based on Reinforcement Learning with Verifiable Rewards (RLVR), 
we propose a novel process-based reward in addition to the standard answer-based and format-based rewards.
In sum, our contributions are as follows:
\begin{itemize}
    \item We propose a RAG-based pipeline that composes image-specific, knowledge-aware reasoning traces by explicitly disentangling visual findings from label- and modality-conditioned domain knowledge.
    \item We introduce an answer-dependent process reward that improves RLVR by encouraging self-consistent and logically plausible reasoning traces.
    \item Experiments on three fundus-image reading benchmarks, \ie FunBench \cite{wei2025funbench}, Omni-Fundus \cite{hu2024omnimedvqa}, and GMAI-Fundus \cite{yegmai} verify the efficacy of \model.   %consistently outperforms multiple strong baselines, showing that 
    A reasoning-enhanced fundus-reading MLLM can indeed be trained using exclusively public data, 94\% of which have image-level labels only.
   \model will be open source.
    %even when the vast majority of available samples provide only image-level labels.
\end{itemize}

\section{Related Work} 
\label{sec:related}

Existing methods for training fundus-reading MLLMs differ not only in the data resources they rely on, but also in how such resources are transformed into supervision and optimized during training. In practice, the form of available data largely determines the downstream learning paradigm.

%\yc{Existing fundus-reading MLLMs differ in their data sources and supervision strategies, with the available data largely determining the learning paradigm.}

One line of work aims for better performance by scaling up training data, typically through the introduction of in-house resources, while retaining relatively coarse supervision. Representative examples include DeepDR-LLM~\cite{DeepDR-LLM} and VisionFM~\cite{qiu2024development}. DeepDR-LLM leverages private real-world clinical supervision together with fundus analysis modules, while VisionFM expands training resources with large-scale public and private ophthalmic images as well as synthetic ophthalmic data for foundation-style visual pretraining. Although effective, such methods mainly rely on larger data volume, and most supervision still remains at the level of images, labels, or coarse clinical associations, without explicitly modeling the reasoning process.

Another line of work seeks to construct richer supervision from raw resources, which naturally leads to a ``data transformation + instruction tuning'' paradigm. With access to private clinical reports or hospital resources, methods such as VisionUnite~\cite{li2024visionunite}, RetinalGPT~\cite{zhu2025retinalgpt}, EyeCareGPT~\cite{li2025eyecaregpt}, and FundusExpert~\cite{fundusexpert} transform annotations, clinical reports, or retinal attributes into richer image--text pairs, dialogues, VQA samples, report-generation data, or structured reasoning-style corpora, and then train ophthalmic MLLMs through multi-stage pretraining or instruction tuning. Compared with plain image-level labels, these methods provide substantially stronger supervision, but such supervision often depends on non-public resources.

% Public-data-only solutions for reasoning-oriented supervision are much more limited. 
Public-data-only reasoning supervision remains rare. OphthaReason~\cite{OphthaReason} is one of the few systems built mainly on public datasets and PMC-derived text. It retrieves electronic clinical reports from PMC to assist the synthesis of reasoning traces, and further incorporates reinforcement learning with answer and format rewards to improve reasoning-oriented generation.
However, because high-quality public clinical text remains scarce, the amount of synthesized reasoning supervision is still limited. Moreover, even with RL, its optimization remains centered on answer correctness and output formatting, without directly supervising the faithfulness or diagnostic validity of the reasoning trace itself.
%\yc{However, scarce public clinical text limits the scale of synthesized supervision, while its RL objective remains focused on answer correctness and formatting rather than the faithfulness or diagnostic validity of the reasoning trace.}

To sum up, existing studies suggest that strong fundus-reading MLLMs usually depend on supervision richer than public image-level labels, while the corresponding learning paradigms are largely determined by how such richer supervision is constructed. Different from the prior research, we aim for 
constructing reasoning traces from image-level labels and consequently exploiting the auto-generated traces for training a reasoning-enhanced model.

%\yc{Process rewards provide additional supervision beyond final-answer correctness by encouraging reliable reasoning traces. 

Prior works on exploiting process-oriented rewards in RLVR focus on math and coding, 
where the quality of reasoning traces can be assessed through formal verification, \eg  formal proof checking \cite{kim2026process}, outcome-based estimation of step contributions \cite{li2026ssvpo}, unit tests \cite{ma2025dynamic}, and execution feedback \cite{fan2026recode,wang2026stepcodereasoner}. These strategies are inapplicable for rewarding reasoning traces in fundus image reading.

\begin{figure*}[t]
\centering
\begin{subfigure}[t]{\linewidth}
    \centering
    \includegraphics[width=\linewidth]{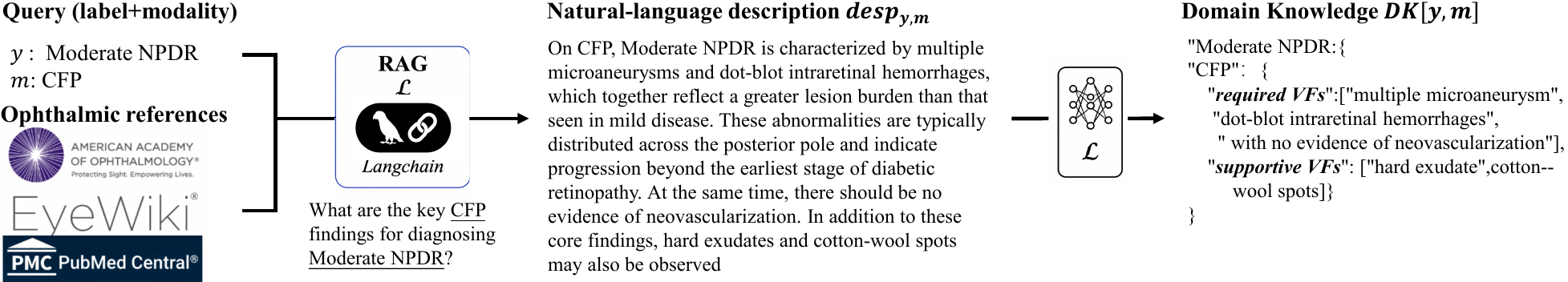}
    \caption{Label-and-modality-specific domain knowledge (\textbf{DK}) acquisition. 
    %A visual-finding (VF) vocabulary is derived by merging the required and supportive VFs stored in $DK$.
    }
    \label{fig:dk}
\end{subfigure}

\begin{subfigure}[t]{0.59\linewidth}
    \centering
    \includegraphics[width=\linewidth]{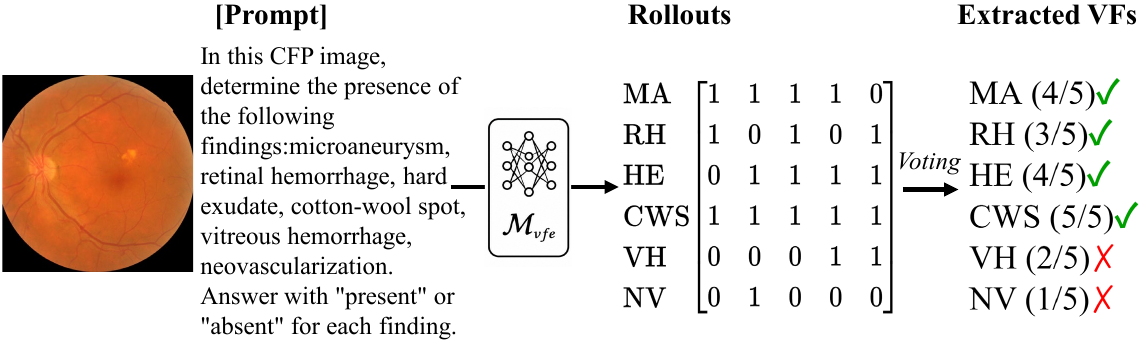}
    \caption{Image-specific visual finding (\textbf{VF}) extraction}
    \label{fig:b}
\end{subfigure}
\hfill
\begin{subfigure}[t]{0.39\linewidth}
    \centering
    \includegraphics[width=\linewidth]{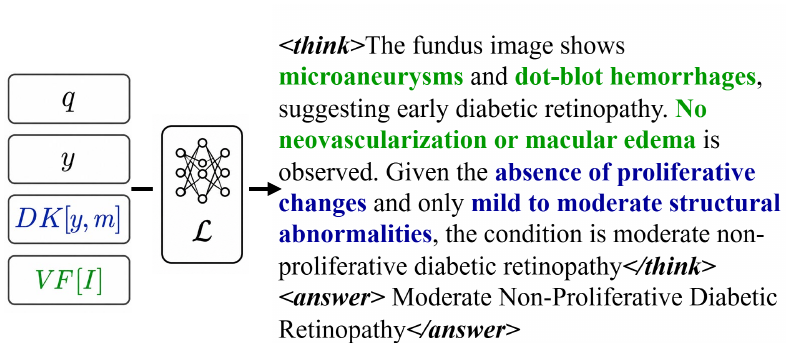}
    \caption{Knowledge-aware reasoning trace composition}
    \label{fig:trace-composition}
\end{subfigure}

\caption{\textbf{Generating visual-finding (VF) embedded reasoning traces} for training \model. 
Given a task label $y$ and fundus modality $m$, we first employ a generic LLM $\mathcal{L}$ (\emph{Qwen3-Max}) to obtain structured label-and-modality-specific domain knowledge $DK[y,m]$ and a VF vocabulary from authoritative ophthalmic references. For each training image $I$ of modality $m$ and labeled with $y$, we extract image-specific VFs  $VF[I]$ by prompting a generic MLLM $\mathcal{M}_{vfe}$ (\emph{Qwen2.5-VL-32B}). Lastly, a reasoning trace is generated for the given image by prompting $\mathcal{L}$ with question $q$, answer $y$, $DK[y,m]$, and $VF[I]$.    
}
\label{fig:method}
\end{figure*}

\section{Proposed \model Solution}\label{sec:method}

\subsection{Problem Statement} \label{ssec:problem}

We aim to develop an MLLM $\mathcal{M}$ that takes a fundus image $I$ and a task-specific question $q$ as input. The objective of $\mathcal{M}$ is to generate a correct answer $\hat{y}$ along with a detailed reasoning trace $\tau$. This process can be expressed more formally as
\begin{equation} \label{eq:general}
    (\hat{y}, \tau) \leftarrow \mathcal{M}(I, q).
\end{equation}

Developing such a model is challenging, as existing public datasets  primarily offer image-level labels without accompanying reasoning traces.
%, see Tab.~\ref{tab:datasets}. 
To address this, we detail in Sec.~\ref{ssec:dataset-curation} our approach to generating these crucial reasoning traces by synergistically combining ophthalmic knowledge with generic LLMs. Subsequently, we elaborate in Sec.~\ref{ssec:post-training} how to effectively exploit these auto-generated reasoning traces for transforming a base model (\emph{Qwen2.5-VL}~\cite{qwen2.5-VL}) into our specialized \model.

\subsection{Adding Reasoning Traces to Existing Data}
\label{ssec:dataset-curation}

Directly prompting an MLLM tends to generate over interpreted and logically inconsistent reasoning traces (\cref{tab:show-traces}). We propose a RAG-based method for generating knowledge-based and image-grounded reasoning traces, see \cref{fig:method}. 

%to generate reasoning traces conditioned on the question, image, and label lead to content that is inconsistent with the image. To address this issue, we construct reasoning traces by explicitly combining two complementary sources: label-and-modality-specifc domain knowledge $DK[y,m]$ and image-specific visual findings $VF[I]$, see Fig.~\ref{fig:method}.

% using general-purpose MLLMs(Qwen3-VL-Plus) 

% The generated rationales often exhibit multiple failure modes, including \textit{over-interpretation}, \textit{missing retinal fidings}, \textit{lack knowledge}, and \textit{inconsistent reasoning chains}. These issues indicate that free-form end-to-end synthesis cannot provide suitable supervision.

\subsubsection{Modality-Specific VF Vocabulary Construction} \label{ssec:dk-vfvocab}

Due to differences in fundus imaging mechanisms, a given lesion may present with distinct visual characteristics across the varied image modalities. For example, \emph{drusens} appear as yellowish-white, round subretinal deposits in CFP, whereas they manifest as mound-like elevations or nodules on OCT. Therefore, for each disease label its VF vocabulary should be constructed in a modality-specific manner. More specifically, given a label $y$ and an image modality $m$, we first acquisite domain knowledge \wrt $y$ and $m$, denoted as $DK[y,m]$, from online ophthalmic references using RAG and a generic LLM $\mathcal{L}$. We then perform a Text-to-Findings operation on $DK[y,m]$ to obtain the vocabulary referred to as $Vocab[y,m]$. We formulate the above process as 
\begin{equation} \label{eq:dk}
\left\{
\begin{array}{ll}
desp_{y,m} & \leftarrow \mathrm{RAG}(y,m),\\
DK[y,m] & \leftarrow \mathcal{L}(desp_{y,m}, \mbox{prompt}), \\
Vocab[y,m] & \leftarrow \mbox{Text-to-Findings}(DK[y,m]).
\end{array}
\right.
\end{equation}

%In particular, the construction is executed in the following three steps: 1) RAG-based domain knowledge acquisition, 2) 

% \cite{AAO}, 
% ~\cite{pmc_homepage}), 
\textbf{RAG-based domain knowledge acquisition}. 
Our RAG module is implemented using Qwen3-Max~\cite{qwen3max} as $\mathcal{L}$ and LangChain \cite{chase2022langchain} for online information retrieval. Given a textual query composed of the label $y$ and the image modality $m$, the RAG module finds web pages relevant \wrt the query from multiple sources (EyeWiki, AAO and PMC), 
downloads the pages, and parses them. Sections describing characteristic symptoms, diagnostic criteria and modality-specific manifestations are retained to form a natural-language description denoted by $desp_{y,m}$. For example, given $y$ as  \emph{Moderate NPDR} and $m$ as \emph{CFP}, the query reads as ``\emph{What are the key \underline{CFP} findings for diagnosing \underline{Moderate NPDR}?}''. The corresponding $desp_{y,m}$ reads as ``\emph{On CFP, Moderate NPDR is characterized by multiple microaneurysms and dot-blot intraretinal hemorrhages, which together reflect a greater lesion burden than that seen in mild disease. These abnormalities are typically distributed across the posterior pole and indicate progression beyond the earliest stage of diabetic retinopathy. At the same time, there should be no evidence of neovascularization. In addition to these core findings, hard exudates and cotton-wool spots may also be observed}''. We refer to the supplement for more examples.
Given $desp_{y,m}$, we prompt $\mathcal{L}$ to act as a knowledge extractor, yielding $DK[y,m]$ as a JSON-style record, see \cref{fig:dk}.
%Fig.1(a).  

%We acquire label-conditioned domain knowledge $DK[y,m]$ through a two-stage pipeline: tool-assisted knowledge collection and structured knowledge canonicalization.

\textbf{VF vocabulary construction}. 
Following LLM extraction, the Text-to-Findings operation applies lightweight regex-based post processing to remove redundant expressions, normalize lexical variants, and consolidate synonymous phrases into their canonical forms.
For instance, in CFP-based diabetic retinopathy (DR) grading, the disease labels are Mild NPDR, Moderate NPDR, Severe NPDR, and PDR. The resultant VF vocabulary contains commonly recognized retinal lesions such as microaneurysm (MA), retinal hemorrhage (RH), hard exudate (HE), cotton-wool spot (CWS), vitreous hemorrhage (VH), and neovascularization (NV). 
Per label and modality, the VFs are stored in two disjoint groups in $DK[y,m]$:  \textbf{required VFs}, which capture core evidence explicitly noted in the retrieved references as characteristic or decisive for the target label, and \textbf{supportive VFs}, which offer auxiliary cues.
%but non-decisive cues.

\subsubsection{Image-specific VF Extraction} \label{ssec:vfe}

Recall that each training image is already associated with a specific label $y$ and a modality label $m$. Using the previously constructed VF vocabulary $Vocab[y,m]$, we tackle the VF extraction task by prompting a generic MLLM, denoted as $\mathcal{M}_{vfe}$, with a set of binary questions, specifically asking whether every entry in $Vocab[y,m]$ is present in the given image. These questions are posed jointly via a customized $\mbox{prompt}_{y,m}$: ``\emph{In this \{m\} image, determine the presence of the following findings: \{Vocab[y,m]\}. Answer with present or absent for each finding}''.
  
To improve the precision of VF extraction, we prompt $\mathcal{M}_{vfe}$ five times to generate a set of five rollouts. These rollouts are then aggregated into a set of predicted VFs, denoted $VF[I]$, by a  majority voting strategy. For each entry in $Vocab[y,m]$, it receives a vote each time it appears in a rollout. An entry is added to $VF[I]$  only if its vote count exceeds two.  Through this rollout-based voting strategy, we extract VFs that $\mathcal{M}_{vfe}$ detects consistently and reliably. The VF extraction process is summarized in Eq. \ref{eq:vfe}.

\begin{equation} \label{eq:vfe}
\left\{
\begin{array}{ll}
\mbox{prompt}_{y,m} & \leftarrow \mbox{Make-Prompt}(m, Vocab[y,m]),\\
\mbox{rollouts} & \leftarrow \mathcal{M}_{vfe}(I, \mbox{prompt}_{y,m}), \\
VF[I] & \leftarrow \mbox{Aggregate}(\mbox{rollouts}).
\end{array}
\right.
\end{equation}

Note that a small portion of our training data (less than 6\%)  have pixel-level lesion annotations. For these images, we directly use their lesion labels as $VF[I]$.

\subsubsection{Knowledge-aware Reasoning Trace Composition} \label{ssec:reasontrace}

Given an image $I$ and its label $y$ \wrt a specific fundus-reading task, we generate a question $q$ by filling out a pre-specified task-specific template with $y$ as the correct answer, following the common practice in previous work  \cite{hu2024omnimedvqa,wei2025funbench}. For the VQA triplet $(I,q,y)$,  we propose to construct a reasoning trace in a knowledge-aware manner, by prompting $\mathcal{L}$ to generate the trace conditioned on both VFs $VF[I]$ and $DK[y,m]$. In particular, $\mathcal{L}$ is guided to (i) summarize the visual findings from $VF[I]$, (ii) connect the findings to the domain knowledge from $DK[y,m]$, and (iii) derive the final answer $y$ accordingly. 
We empirically observe that compared to free-form chain-of-thought generation, the above prompting strategy produces reasoning traces that are more structured and easier to verify, see  \cref{tab:show-traces}.

 %, but only narrows the space of disease-relevant VFs, whose presence must still be verified in the image. The corresponding $DK[y,m]$ then establishes the medical links between the verified VFs and $y$.

Using the pipeline illustrated in \cref{fig:method}, we initially generated 146,425 reasoning traces in total. 
A quality check was then automatically performed by prompting $\mathcal{L}$ to identify problematic traces, including modality inconsistency, omission of required VFs, inclusion of redundant or incorrect VFs, and omission or mismatch of domain knowledge, \etc 
Further details are provided in the supplement. 
%Ultimately, 80,115 traces were retained and added to the training data for MLLM post-training.
Ultimately, 80,115 traces were retained for MLLM post-training.
Since the traces are generated in a reverse-engineering manner, they inevitably contain errors. We therefore use the traces only as a form of weak intermediate supervision. This approach is inline with \cite{jiang-etal-2025-wise} that leverages synthesized explanations for improving LLM-based image classification.

%\yc{It is worth pointing out that the label $y$ alone does not determine the reasoning trace. Each of the VFs defined in  $DK[y,m]$ is extracted based on the image content, see \cref{fig:b}.  \xr{to rewrite} We therefore regard the resulting traces as a weak intermediate supervision, rather than unique ground-truth reasoning paths, consistent with prior work using weakly supervised synthesized explanations for multimodal image classification~\cite{jiang-etal-2025-wise}.}

\subsection{Improving RLVR with a Process Reward} 
\label{ssec:post-training}

With the reasoning-trace-enriched training data, we now proceed to post-train a generic model to its fundus-reading counterpart $\mathcal{M}$. We adopt the widely used RLVR algorithm based on  Group Relative Policy Optimization (GRPO) \cite{grpo}. Given a specific VQA triplet $(I, q, y)$ as a training example, the basic idea of GRPO is to first let $\mathcal{M}$ generate a group of $G$ different outputs,  commonly known as rollouts, which are denoted by $\{(\hat{y}_{i}, \tau_{i})\}_{i=1}^G$, where $\hat{y}_{i}$ is the predicted answer in the rollout $i$ and $\tau_i$ as the generated reasoning trace. Based on $\hat{y}_{i}$, $\tau_i$ and the input, a reward value $r_i$ for the rollout $i$ is calculated. The average reward of the group is then used as a baseline to calculate a relative advantage for each rollout. The model is optimized by maximizing these relative advantages. The calculation of $r_i$ is thus critical. 

\begin{figure}[!htbp]
    \centering
    \begin{subfigure}[t]{\columnwidth}
        \centering
        \includegraphics[width=\linewidth]{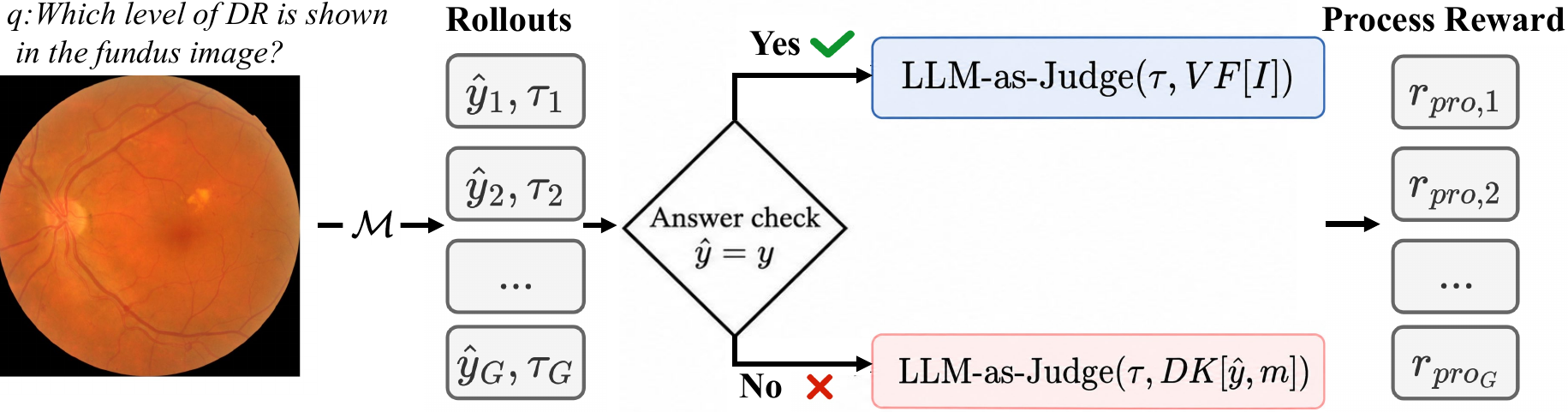}
        \caption{Answer-dependent process reward computation.}
        \label{fig:calc-rpro}
    \end{subfigure}
    % \hfill
    \vspace{1em}
    \begin{subfigure}[t]{\columnwidth}
        \centering
        \includegraphics[width=\linewidth]{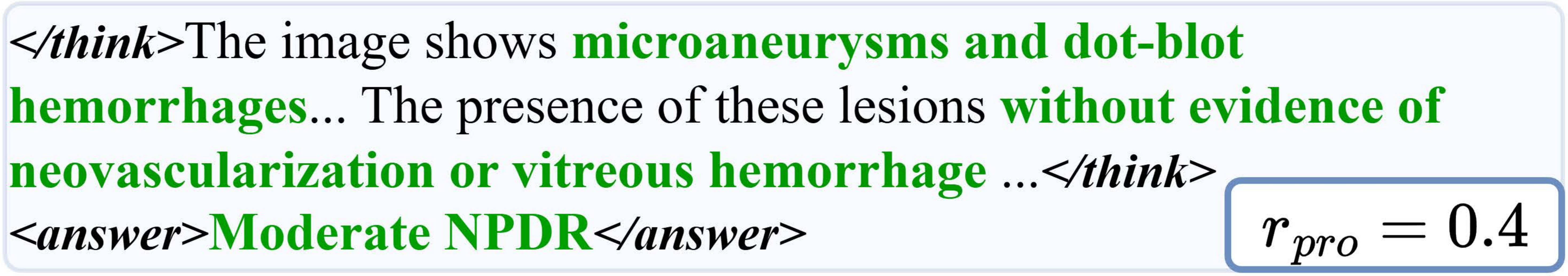}
        \caption{
        %A rollout with a correct answer.
        Correct rollout with a logically \emph{plausible} reasoning trace. 
        }
        \label{fig:correct-plausible}
    \end{subfigure}
    % \hfill
    \vspace{0.83em}
    \begin{subfigure}[t]{\columnwidth}
        \centering
        \includegraphics[width=\linewidth]{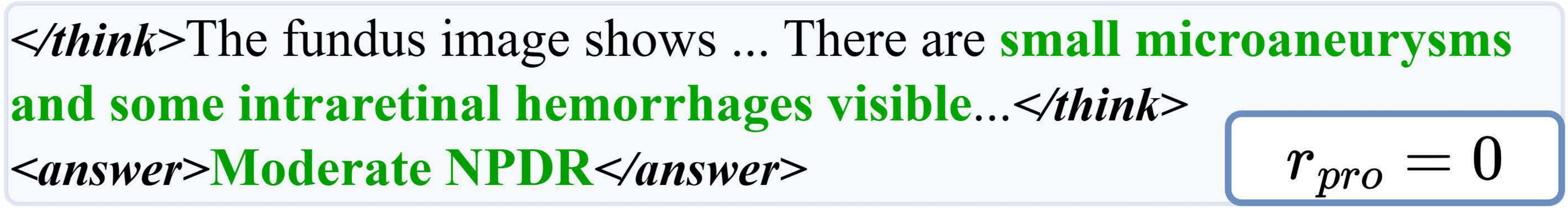}
        \caption{
        Correct rollout  with a logically \emph{tenuous} reasoning trace.
        %A rollout with an incorrect answer.
        }
        \label{fig:correct-tenuous}
    \end{subfigure}
    % \hfill
    \vspace{0.83em}
    \begin{subfigure}[t]{\columnwidth}
        \centering
        \includegraphics[width=\linewidth]{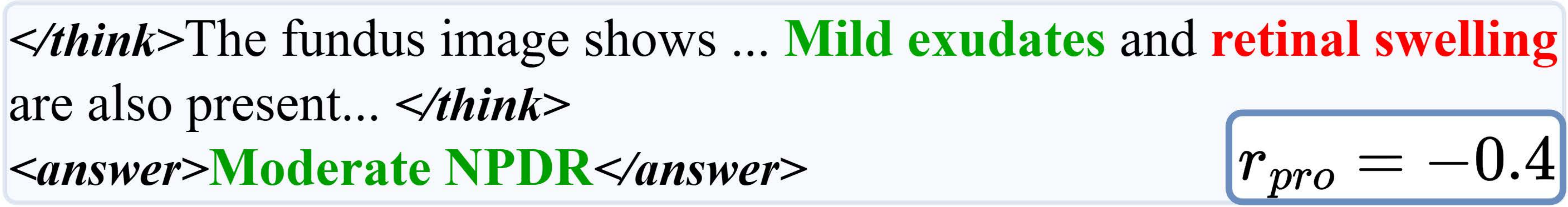}
        \caption{
        Correct rollout with a logically \emph{flawed} reasoning trace.
        %A rollout with an incorrect answer.
        }
        \label{fig:correct-flawed}
    \end{subfigure}
    % \hfill
    \vspace{0.83em}
    \begin{subfigure}[t]{\columnwidth}
        \centering
        \includegraphics[width=\linewidth]{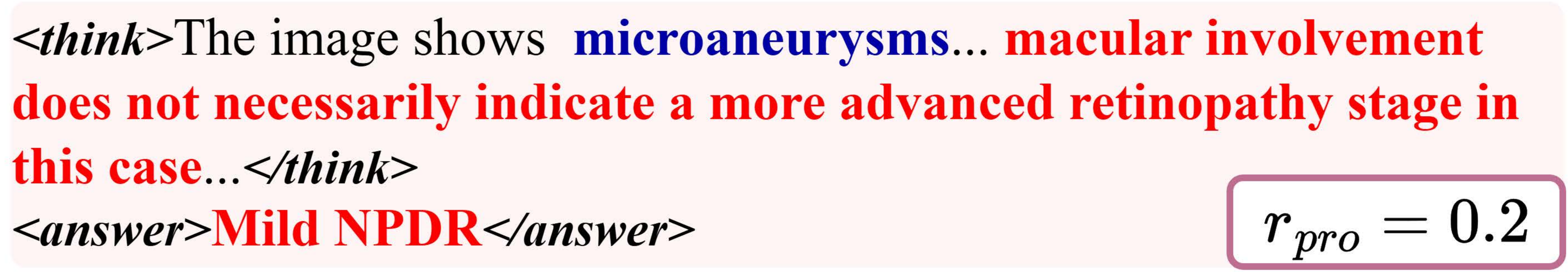}
        \caption{
         \emph{Incorrect} rollout  with a logically \emph{plausible} reasoning trace.
        %A rollout with an incorrect answer.
        }
        \label{fig:incorrect-plausible}
    \end{subfigure}
    % \hfill
    \vspace{0.83em}
    \begin{subfigure}[t]{\columnwidth}
        \centering
        \includegraphics[width=\linewidth]{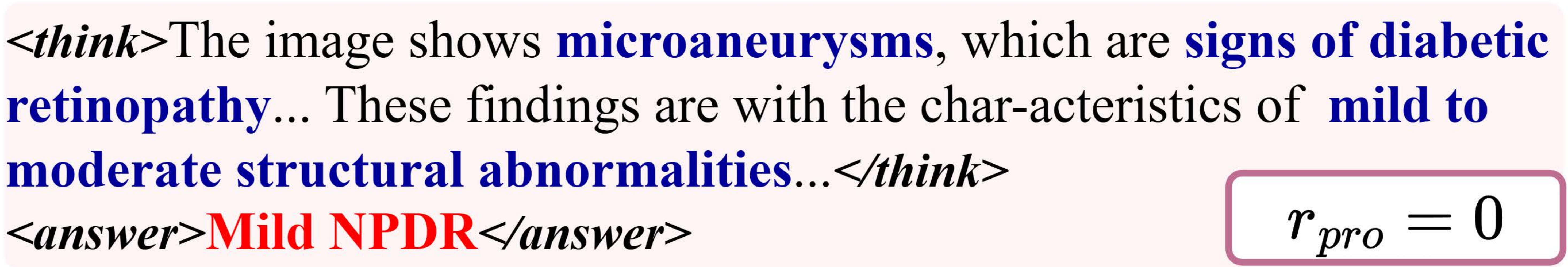}
        \caption{
        \emph{Incorrect} rollout  with a logically \emph{flawed} reasoning trace.
        %A rollout with an incorrect answer.
        }
        \label{fig:incorrect-flawed}
    \end{subfigure}
    \caption{
    \textbf{Our process reward} $r_{pro}$. Best viewed digitally.
    }
    \label{fig:rpro}
\end{figure}

Prior work, such as OphthaReason \cite{OphthaReason}, determines the reward based solely on whether the predicted answer $\hat{y}_i$ is correct and whether the generated trace $\tau$ adheres to the required format. Specifically, the reward is computed as the sum of a binary answer-based reward $r_{ans,i}$ and a binary format-based reward $r_{fmt,i}$, thereby completely disregarding the quality of $\tau$ itself.  By contrast, we explicitly account for trace quality by introducing a process reward $r_{pro,i}$. 
% Prior work, such as OphthaReason~\cite{OphthaReason}, rewards each rollout only according to answer correctness and format compliance. Specifically, its reward is the sum of a binary answer reward $r_{ans,i}$ and a binary format reward $r_{fmt,i}$, without explicitly evaluating the quality of the generated reasoning trace $\tau_i$. 
Our design of $r_{pro,i}$ is guided by the following considerations. Ideally, we want $\mathcal{M}$ to produce a correct answer based on correct visual findings. Therefore, when $\hat{y}_i$ matches $y$, $r_{pro,i}$ should verify that $\tau$ aligns with $VF[I]$, \ie the visual findings previously extracted from the training image $I$. However, such a criterion might be over strict, especially in the early stages of training when the model has not yet learned to answer correctly. To address this, even when $\hat{y}_i$ is incorrect, the model can still receive a reward if its reasoning process remains  logically plausible \wrt $DK[\hat{y}_i,m]$, \ie the domain knowledge associated with the incorrect answer. Taking these into account, we propose to compute $r_{i,pro}$ in an answer-dependent manner as follows:
\begin{equation} \label{eq:rpro}
r_{\mathrm{pro},i} =
\begin{cases}
\text{LLM-as-judge}(\tau_{i}, VF[I]), & \hat{y}_{i}=y,\\[4pt]
\text{LLM-as-judge}(\tau_{i}, DK[\hat{y}_{i},m]), & \hat{y}_{i}\neq y,
\end{cases}
\end{equation}
% where $\mbox{LLM-as-judge}$ returns a value by prompting $\mathcal{L}$ to assess $\tau_i$ against the provided reference, which is either $VF[I]$ or $DK[\hat{y}_{i},m])$. In particular, when $\hat{y}_i$ is correct, $\mathcal{L}$ is instructed to rate the trace  as \emph{plausible}, \emph{tenuous}, or \emph{flawed}, corresponding to reward values of $0.4$, $0$ and $-0.4$, respectively, see \cref{fig:rpro}. When $\hat{y}_i$ is incorrect, $\mathcal{L}$ evaluates whether the trace is logically plausible, yielding a reward value of $0.2$ if plausible and $0$ otherwise. Note that summing the two terms in Eq. \ref{eq:rpro} is prone to reward hacking, resulting in suboptimal performance. 
% The overall reward $r_i$ is obtained by summing $r_{ans,i}$, $r_{fmt,i}$ and $r_{pro,i}$.
where $\mbox{LLM-as-judge}$, by prompting $\mathcal{L}$, assess $\tau_i$ against the provided reference, which is either $VF[I]$ or $DK[\hat{y}_{i},m]$. In particular, when $\hat{y}_i$ is correct, $\mathcal{L}$ is instructed to rate the trace as \emph{plausible}, \emph{tenuous}, or \emph{flawed}, corresponding to reward values of $0.4$, $0$ and $-0.4$, respectively, see \cref{fig:rpro}. When $\hat{y}_i$ is incorrect, $\mathcal{L}$ evaluates whether the trace is logically plausible, yielding a reward value of $0.2$ if plausible and $0$ otherwise. These reward scales are empirically determined to ensure the trace reward being substantially lower than the answer counterpart. As such, the answer reward provides  primary supervision, while the trace reward is mainly for injecting ophthalmic knowledge.
%and thus mitigating errors introduced during VF extraction.} 
%Note that summing the two terms in Eq. \ref{eq:rpro} is prone to reward hacking, resulting in suboptimal performance.
The overall reward $r_i$ is obtained by summing $r_{ans,i}$, $r_{fmt,i}$ and $r_{pro,i}$.

To assess the reliability of LLM-as-judge, a small-scale manual evaluation was conducted. We constructed a class-balanced testset, by manually collecting 100 plausible rollouts and 100 flawed rollouts, both with fluent tones. The LLM (Qwen3-Max) has an accuracy of 0.79 (\cref{tab:llm_judge_reliability}).
% \begin{table}[!htbp]
% \centering
% \caption{Manual assessment of LLM-as-judge used for the process reward computation.}
% \label{tab:llm_judge_reliability}
% \small
% \begin{tabular}{lcc}
% \toprule
% & \textbf{Human Plausible} & \textbf{Human Flawed} \\
% \midrule
% \textbf{LLM Plausible} & 77 & 19 \\
% \textbf{LLM Flawed}    & 23 & 81 \\
% \bottomrule
% \end{tabular}
% \end{table}

\begin{table}[!htbp]
\centering
\caption{\textbf{Manual assessment of LLM-as-judge}.}
%used for the process reward computation.}
\label{tab:llm_judge_reliability}
\footnotesize
\setlength{\tabcolsep}{4pt}
\renewcommand{\arraystretch}{0.9}
\begin{tabular}{lcc}
\toprule
\textbf{LLM Judgment} & \textbf{Human Plausible} & \textbf{Human Flawed} \\
\midrule
Plausible & 77 & 19 \\
Flawed    & 23 & 81 \\
\bottomrule
\end{tabular}
\end{table}

\subsection{Two-Stage Training} 
\label{ssec:training}
Note that directly post-training the base model using RLVR is problematic, as rollouts in the early stages of training are predominantly incorrect and the generated reasoning traces often vary substantially in length and format. As a consequence, reward values tend to be near zero, making RLVR slow to boot up. We therefore employ a two-stage training procedure. In the first stage, SFT is performed to adapt the model to the various fundus image reading tasks. We then refine the model in the second stage using the process-reward-enhanced RLVR.
\section{Experiments} 
\label{sec:eval}
\subsection{Experimental Setup} \label{ssec:eval-setup}
\textbf{Test sets}.  We adopt three public test sets: FunBench \cite{wei2025funbench}, OmniMedVQA \cite{hu2024omnimedvqa}, and GMAI-MMBench \cite{yegmai}, see  \cref{tab:testsets}. FunBench evaluates an MLLM's fundus reading skills via a hierarchical task organization across four levels, including \emph{modality perception} (L1), \emph{anatomy perception} (L2), \emph{lesion analysis} (L3), and \emph{disease diagnosis} (L4). For OmniMedVQA and GMAI-MMBench, we adopt their subsets related to fundus image reading. In particular, the following is selected from OmniMedVQA: \emph{modality recognition} (MR), \emph{anatomy identification} (AI), \emph{lesion grading} (LG), and \emph{disease diagnosis} (DD). As for GMAI-MMBench, we use \emph{attribute recognition} (AR), \emph{blood vessel recognition} (BVR), \emph{disease diagnosis} (DD), and \emph{severity grading} (SG).
Accordingly, 
%For the ease of reference, 
we rename the two subsets as \textbf{Omni-Fundus} and \textbf{GMAI-Fundus}, respectively. 
%Tab.~\ref{tab:testsets} shows the basic statistics of the three test sets. 

Data contamination was carefully avoided. Pictorial content-based matching was performed to detect and consequently remove any overlap between training and test images. Neither test data nor its metadata was used for prompt construction or trace generation.

\begin{table}[htbp!]
\caption{\textbf{Three test sets used in our experiments}.}
\label{tab:testsets}
\setlength{\tabcolsep}{2pt}
\renewcommand{\arraystretch}{1}
\centering
\resizebox{0.9\linewidth}{!}{%
\begin{tabular}{@{}l r r l@{}}
\toprule
\textbf{Test set} & \textbf{Modalities} (\#Images) & \textbf{\#VQA triplets} \\
\midrule
FunBench
& CFP (7,608) / OCT (6,076) / SLO (2,664) 
& 91,810 
\\
Omni-Fundus  
& CFP (6,875) / OCT (2,319) 
& 9,194
\\
GMAI-Fundus
& CFP (1330) / OCT (183) /SLO(47)
& 1,560
\\
\bottomrule
\end{tabular}}
\end{table}

\begin{table*}[!htbp]
\centering
\caption{
\textbf{Performance of varied training setups}.
\textbf{Bold} and \underline{underline} denote the best and second-best, respectively.
}
\label{tab:main_ablation_all_benchmarks}
\setlength{\tabcolsep}{2pt}
\renewcommand{\arraystretch}{1}
\resizebox{0.9\textwidth}{!}{
\begin{tabular}{@{}c l
cccc c
cccc c
cccc c
cccc@{}}
\toprule

\multirow{2}{*}{\textbf{\#}}
& \multirow{2}{*}{\textbf{Training Setup}}
& \multicolumn{4}{c}{\textbf{Overall}} &
& \multicolumn{4}{c}{\textbf{FunBench}} &
& \multicolumn{4}{c}{\textbf{Omni-Fundus}} &
& \multicolumn{4}{c}{\textbf{GMAI-Fundus}} \\

\cmidrule{3-6} \cmidrule{8-11} \cmidrule{13-16} \cmidrule{18-21}

& & \textit{Avg} & \textit{FunBench} & \textit{Omni-Fundus} & \textit{GMAI-Fundus} &
& \textit{L1} & \textit{L2} & \textit{L3} & \textit{L4} &
& \textit{MR} & \textit{AI} & \textit{LG} & \textit{DD} &
& \textit{AR} & \textit{BVR} & \textit{DD} & \textit{SG} \\

\midrule

\textbf{A}:
& Base (Qwen2.5-VL-3B)
& 49.3 & 41.0 & 66.4 & 40.4 &
& 88.1 & 40.8 & 21.7 & 13.6 &
& 96.4 & 87.4 & 44.6 & 37.3 &
& 31.3  & 74.7 & 37.1 & 18.6 \\

\multicolumn{21}{@{}l}{\textit{Training without reasoning trace:}} \\

\textbf{B}:
& \textbf{A} + SFT
& 55.1 & 58.9 & 71.1 & 35.3 &
& 69.3 & \textbf{95.8} & 20.4 & 50.0 &
& 94.2 & 91.9 & 51.7 & 46.4 &
& 24.7  & 65.3 & 33.4 & 17.6 \\

\textbf{C}:
& \textbf{A} + GRPO
& 57.0 & 53.9 & 68.1 & 48.9 &
& 76.8 & 71.5 & 27.2 & 40.2 &
& 92.4 & 92.8 & 36.2 & 50.9 &
& \underline{38.7}  & 85.3 & 32.0 & 39.7 \\

\textbf{D}:
& \textbf{A} + SFT + GRPO
& 53.6 & 53.1 & 71.3 & 36.4 &
& 88.4 & 57.7 & 30.5 & 35.7 &
& 95.4 & \underline{93.8} & 43.4 & 52.7 &
& 28.0  & 37.3 & 43.3 & 37.1 \\

\multicolumn{21}{@{}l}{\textit{Training with reasoning traces:}} \\

\textbf{E}:
& \textbf{A} + SFT
& 59.0 & 61.9 & 67.7 & 47.2 &
& 90.4 & 83.8 & 18.1 & \textbf{55.4} &
& \textbf{97.6} & 89.3 & 26.2 & 57.7 &
& 38.0  & 81.3 & 41.2 & 28.5 \\

\textbf{F}:
& \textbf{A} + SFT + GRPO
& 60.4 & 64.7 & 71.4 & 45.0&
& 92.1 & 94.3 & 29.8 & 42.7 &
& 92.7 & 87.2 & 53.7 & 51.8 &
& 22.0  & 72.0 & 51.0 & 35.2 \\

\rowcolor{blue!20}
\textbf{G}:
& \model-3B & \textbf{68.4} & \textbf{67.1} & \textbf{79.8} & \textbf{58.5}
& & 90.5 & 90.9 & \textbf{33.9} & \underline{52.9} &
& 93.8 & 92.1 & \textbf{63.4} & \textbf{69.7} &
& 38.0  & \textbf{92.0} & \textbf{57.0} & \underline{46.9} \\

\multicolumn{21}{@{}l}{\textit{Ablation on the process reward:}} \\

\textbf{H}:
& Using the VF item only
& 62.6 & 63.2 & 75.6 & 49.1 &
& 91.0 & 89.7 & 30.7 & 41.6 &
& 93.2 & 91.7 & 54.5 & \underline{63.0} &
& 31.3  & 69.3 & 51.5 & 44.2 \\

\textbf{I}:
& Using the DK item only
& \underline{66.8} & \underline{66.6} & \underline{77.4} &  \underline{56.5} &
& \underline{92.8} & \underline{94.5} & 31.4 & 47.9 &
& \underline{97.0} & \underline{93.8} & \underline{58.4} & 60.3 &
& \textbf{40.7} & \underline{90.7} & \underline{52.9} & 41.7 \\

\textbf{J}:
& Summing the VF and DK items
& 64.7 & 61.7 & 76.7 &55.7 &
& \textbf{93.4} & 82.6 & \underline{31.7} & 39.1 &
& 94.3 & \textbf{96.0} & 57.1 & 59.5 &
& 35.3 & 88.0 & 52.6 & \textbf{47.1} \\

\bottomrule
\end{tabular}
}

\end{table*}

\textbf{Performance metrics}. For each test set, we report its official metric: F1-score for FunBench, and accuracy for Omni-Fundus and GMAI-Fundus.

We refer to the supplement for training details.

\subsection{Training \emph{w/} or \emph{w/o} Reasoning}
\label{sec:abreasoning}

\textbf{The need of reasoning traces}. 
We first evaluate whether enriching the training data with the generated reasoning traces is necessary.
Tab.~\ref{tab:main_ablation_all_benchmarks} shows the performance of training setups without reasoning traces (Setup B--D) and with reasoning traces (Setup E--G). The clearly better performance of the latter group confirms the necessity of incorporating reasoning traces into the standard VQA-triplet based training data.

\textbf{The need of process-based supervision}. Let us take a closer look at Setup B, which adapts the base model by SFT. Its overall improvement (49.3 $\rightarrow$ 55.1) is largely contributed by a substantial gain on FunBench (41.0 $\rightarrow$ 58.9), although with a clear decrease on GMAI-Fundus  (40.4 $\rightarrow$ 35.3). %Moreover, 
Combining SFT and GRPO without reasoning trace (Setup D) does not yield further improvement. Rather, the overall score drops to 53.6.
These results suggest that answer-only supervision is insufficient for reliably initializing reasoning-oriented RLVR, and that simply stacking SFT and RL without process-based supervision may even be detrimental.

When reasoning traces are provided, SFT + GRPO (Setup F) clearly outperforms its trace-free counterpart (Setup D), from 53.6 to 60.4. However, the lower performance of Setup F relative to Setup E on certain columns, for instance, \emph{L4} in FunBench (55.4 $\rightarrow$ 42.7) and \emph{AR} in GMAI-Fundus (38.0 $\rightarrow$ 22.0), suggests that not all earlier gains from SFT are consistently preserved. Hence, while reasoning traces make RLVR more promising, relying on the answer and format-based rewards is inadequate to unleash the value of traces.

\textbf{Evaluating VF extraction}. We evaluate VF extraction on Retinal-Lesions \cite{retinal-lesions} that has ground truth available for multiple DR-related lesions. As a baseline, we prompt Qwen3-VL-Plus, which is stronger than Qwen2.5-VL-32B used in our VF extraction pipeline, to generate CoT traces given the VQA triplets. Corresponding VFs are then extracted from the generated trace.
Our method is clearly better (46.8 $\rightarrow$ 62.6), though much room remains for improvement (\cref{tab:lesion_se_sp_s2}). Despite the imperfections in the extracted VFs, they are embedded in the reasoning traces in a knowledge-aware manner, rendering  traces valuable for injecting ophthalmic knowledge into the MLLM.
%via post-training.

\begin{table}[t]
\centering
\caption{\textbf{Evaluation of VF extraction}.
%Test set: Retinal-Lesions \cite{retinal-lesions}. 
Performance metrics: Sensitivity, Specificity and their harmonic mean (S2).}
\label{tab:lesion_se_sp_s2}
\setlength{\tabcolsep}{3pt} 
\renewcommand{\arraystretch}{1}

\resizebox{0.9\linewidth}{!}{
\begin{tabular}{@{}l *{3}{>{\raggedleft\arraybackslash}p{2.4em}} c *{3}{>{\raggedleft\arraybackslash}p{2.4em}}@{}}
\toprule
\multirow{2}{*}{\textbf{VF}}
& \multicolumn{3}{c}{\textbf{Ours}} &
& \multicolumn{3}{c}{\textbf{Qwen3-VL-Plus}} \\
\cmidrule{2-4} \cmidrule{6-8}
& \textit{Sen.} & \textit{Spe.} & \textit{S2} &
& \textit{Sen.} & \textit{Spe.} & \textit{S2} \\
\midrule
Microaneurysm (MA)       & 78.3 & 49.5 & 60.7 & & 58.4 & 23.3 & 33.3 \\
Retinal hemorrhage (RH)  & 73.7 & 44.4 & 55.4 & & 45.5 & 51.0 & 48.1 \\
Hard exudate (HE)        & 78.7 & 58.1 & 66.9 & & 49.9 & 65.8 & 56.7 \\
Cotton-wool spot (CWS)  & 48.2 & 94.4 & 63.8 & & 31.9 & 85.3 & 46.4 \\
Vitreous hemorrhage (VH)  & 62.5 & 75.4 & 68.3 & & 37.5 & 81.0 & 51.3 \\
Neovascularization (NV)   & 55.8 & 66.3 & 60.6 & & 32.6 & 72.2 & 44.9 \\
\midrule
Mean & 66.2 & 64.7 & \textbf{62.6} & & 42.6 & 63.1 & 46.8 \\
\bottomrule
\end{tabular}
}
\end{table}

\textbf{Evaluating reasoning traces}. 
We conduct an expert evaluation on both synthesized traces and inference-time traces. 
Two sets of 110 CFP images were randomly sampled from the training and test sets, respectively.
Six ophthalmologists were
recruited to collectively rate the traces on clinical correctness and
visual faithfulness using a 5-point Likert scale (higher is better). The experts were also asked to identify any hallucinated VFs in the traces generated by free-form CoT and our pipeline, respectively. Per image, they indicated which trace they deemed better or if the two were comparable. To avoid bias, the order of the two traces was randomly swapped before presenting to the annotators. 
The evaluation clearly shows the superiority of our
generated traces (\cref{tab:expert_audit}).
Compared to its counterpart post-trained without using the synthesized traces (Setup C, Tab.~\ref{tab:main_ablation_all_benchmarks}), \model exhibits greater visual-textual consistency and lower hallucination.

%Nonetheless, a hallucination rate of 25.6\%  remains high for real-world use. 

% This may stem from predominantly disease-level supervision, which does not explicitly verify individual visual findings, and training objectives that prioritize diagnostic correctness and plausible reasoning over the visual faithfulness of these findings. Moreover, our strict trace-level evaluation considers a trace hallucinated if it contains even one unsupported finding. Thus, this rate should not be interpreted as the proportion of incorrect diagnoses or individual visual findings.
%a substantial performance gap remains between our research prototype and a real-world product.
% \yc{We also constructed a class-balanced test set comprising 100 medically faithful and 100 medically flawed yet fluent rollouts. Qwen3-Max achieved an accuracy of 0.79 on this set.}

\begin{table}[t]
\centering
\caption{\textbf{Expert evaluation of the reasoning-trace quality}.}
\label{tab:expert_audit}
\setlength{\tabcolsep}{2.5pt}
\renewcommand{\arraystretch}{0.92}

\resizebox{0.95\columnwidth}{!}{
\begin{tabular}{@{}lrrrr@{}}
\toprule
\textbf{Reasoning-Trace Source} 
& \textbf{Clinical}$\uparrow$
& \textbf{Faith}.$\uparrow$

& \textbf{Prefer}$\uparrow$
& \textbf{Hallucinated lesions}$\downarrow$\\
\midrule
\multicolumn{5}{@{}l}{\textit{Trace synthesis}:} \\
Free-form CoT (Qwen3-VL-Plus)
& 3.7 & 1.8 & 9.5\% & 54.8\%  \\
\emph{Our pipeline}
& \textbf{4.4} & \textbf{3.9} & \textbf{82.9}\% & \textbf{24.1}\%  \\
\midrule
\multicolumn{5}{@{}l}{\textit{Inference-time trace generation}:} \\
Post-trained \emph{w/o} reasoning traces
& 3.6 & 2.2 & 11.4\%  & 49.4\% \\
\emph{\model-3B}
& \textbf{4.2} & \textbf{3.3} & \textbf{64.8}\%  & \textbf{25.6}\% \\
\bottomrule
\end{tabular}
}
\end{table}

\subsection{Ablation on the Process Reward}

Since the only difference between Setup G and Setup F is the use of $r_{pro}$, the superior performance of Setup G over Setup F (60.4 $\rightarrow$ 68.4) verifies the effectiveness of the process reward. Gains are particularly evident on diagnosis-oriented tasks, including FunBench \emph{L4} (42.7 $\rightarrow$ 52.9), Omni-Fundus \emph{LG} (53.7 $\rightarrow$ 63.4) and \emph{DD} (51.8 $\rightarrow$ 69.7), as well as GMAI-Fundus \emph{BVR} (72.0 $\rightarrow$ 92.0), \emph{DD} (51.0 $\rightarrow$ 57.0), and \emph{SG} (35.2 $\rightarrow$ 46.9). In contrast, for low-level perception tasks such as FunBench \emph{L1}/\emph{L2}, Setup G performs slightly worse.  These results suggest that the process reward primarily benefits fundus reading tasks that require high-level reasoning.

\begin{figure}[!t]
\centering
\begin{subfigure}[]{\linewidth}
\centering
\includegraphics[width=0.9\linewidth]{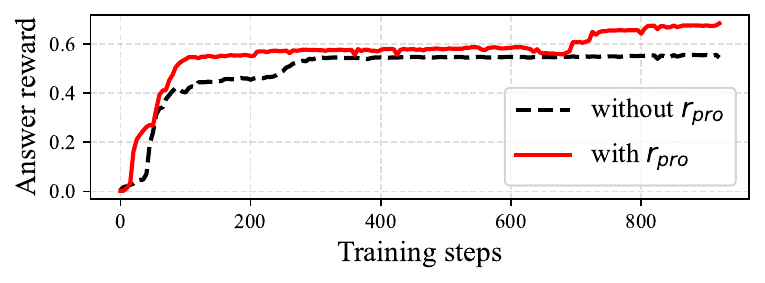}
\caption{The influence of $r_{pro}$ on the answer reward.}
\label{fig:reward_curve}
\end{subfigure}

\begin{subfigure}[]{\linewidth}
\centering
\includegraphics[width=0.9\linewidth]{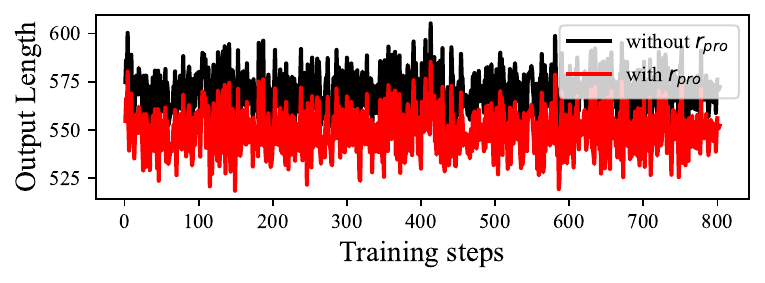}
\caption{The influence of $r_{pro}$ on the output length of $\mathcal{M}$}
\label{fig:rollout_length}
\end{subfigure}
\caption{\textbf{The influence of the process reward $r_{pro}$}.  \model achieves larger answer rewards with shorter output length.
}  
% (\textbf{a})~Faster convergence and higher steady-state performance with context. 
% (\textbf{b})~Shorter reasoning traces with context. Best viewed in color. }
\label{fig:training_dynamics}
\end{figure}

Our ablation on the process reward (Setup H--J,   \cref{tab:main_ablation_all_benchmarks}) shows that the DK item is more effective than its VF counterpart. Summing the two items (Setup J) instead of using them selectively (Eq. \ref{eq:rpro}) is suboptimal. 
\cref{fig:training_dynamics} further shows the benefit of the process reward: making the model achieve higher answer rewards with shorter and hence more concise rollouts during RLVR training.

\subsection{\model \emph{versus} Others}
\label{ssec:vs-others}

To demonstrate the challenging nature of fundus image reading, we report the performance of various public 3B/7B MLLMs on the three test sets. In addition to Qwen2.5-VL, which serves as our base model, we include InternVL2.5-8B \cite{chen2024expanding} as another generic MLLM. For SFT-based medical MLLMs, we include  HealthGPT-M3-7B \cite{lin2025healthgptmedicallargevisionlanguage}, HuatuoGPT-Vision-7B \cite{chen2024huatuogpt}, and Lingshu-7B \cite{xu2025lingshu}. For RLVR-based medical MLLMs, we select MedVLM-R1 \cite{pan2025medvlm} and QoQ-Med \cite{dai2025qoq}. As for fundus-reading MLLMs, we include Med-R1-fundus \cite{med-r1}, FundusExpert-8B \cite{fundusexpert} and OphthaReason \cite{OphthaReason}.

\begin{table}[!htbp]
\centering
\caption{\textbf{Comparison with other MLLMs}.
}
\label{tab:main_arch_perf}
% \scriptsize
\setlength{\tabcolsep}{2.5pt}
\renewcommand{\arraystretch}{1}

\resizebox{\columnwidth}{!}{
\begin{tabular}{@{}lrrrr@{}}
\toprule
\textbf{Model}
& \textbf{FunBench}
& \textbf{\makecell{Omni-Fundus}}
& \textbf{\makecell{GMAI-Fundus}}
& \textbf{Avg.} \\
\midrule

\multicolumn{5}{@{}l}{\textbf{Generic MLLMs}} \\
Qwen2.5-VL-3B~\cite{qwen2.5-VL}
& 41.0 & 66.4 & 40.4 & 49.3 \\
Qwen2.5-VL-7B~\cite{qwen2.5-VL}
& 51.0 & 55.6 & 52.4 & 53.0 \\
InternVL2.5-8B~\cite{chen2024expanding}
& 46.1 & 68.1 & 45.2 & 53.1 \\

\multicolumn{5}{@{}l}{\textbf{Medical MLLMs with SFT}} \\
HealthGPT-M3-7B~\cite{lin2025healthgptmedicallargevisionlanguage}
& 52.4 & 64.0 & 50.3 & 55.6 \\
HuatuoGPT-Vision-7B~\cite{chen2024huatuogpt}
& 59.0 & 69.6 & 51.4 & 60.0 \\
Lingshu-7B~\cite{xu2025lingshu}
& 55.1 & \textbf{97.7} & 62.7 & \underline{71.8} \\

\multicolumn{5}{@{}l}{\textbf{Medical MLLMs with RLVR}} \\
MedVLM-R1~\cite{pan2025medvlm}
& 46.2 & 56.3 & 40.0 & 47.5 \\
QoQ-Med-7B~\cite{dai2025qoq}
& 62.5 & 57.9 & 48.0 & 56.2 \\

\multicolumn{5}{@{}l}{\textbf{Fundus-reading MLLMs with SFT}} \\
FundusExpert-8B~\cite{fundusexpert}
& 48.3 & 78.4 & \textbf{76.9} & 67.9 \\

\multicolumn{5}{@{}l}{\textbf{Fundus-reading MLLMs with RLVR}} \\
OphthaReason-Intern~\cite{OphthaReason}
& 44.9 & 57.7 & 47.9 & 50.2 \\
OphthaReason-Qwen~\cite{OphthaReason}
& 48.8 & 62.4 & 53.0 & 54.7 \\
Med-R1-fundus~\cite{med-r1}
& 48.2 & 79.2 & 37.2 & 54.9 \\

%\midrule
\rowcolor{blue!20}
\textbf{\model{}-3B}
& \underline{67.1} & 79.8 & 58.5 & 68.4 \\
\rowcolor{blue!20}
\textbf{\model{}-7B}
& \textbf{69.2} & \underline{91.1} & \underline{66.1} & \textbf{75.5} \\

\bottomrule
\end{tabular}
}

\end{table}

As \cref{tab:main_arch_perf} shows, \model compares favorably against the others. Nevertheless, since the specialized models are post-trained under varied setups, the conclusion is drawn more at a solution level, other than at an ingredient level. 
Note that LingShu’s training data includes Omni-Fundus \cite{xu2025lingshu}, so the model performs nearly perfectly on this dataset.
Additionally, we compare the MLLMs with pure-vision baselines that perform lesion and disease recognition by linear probing of 
% generic or ophthalmic 
vision foundation models %\cite{zhou2023foundation,sun2025dataefficient,oquab2024dinov2} 
(Tab.~\ref{tab:purevisionpart}).
\model remains highly competitive.

\begin{table}[t]
\centering
\caption{\textbf{MLLMs \emph{versus} pure-vision baselines}. Metric: F1.}
\label{tab:purevisionpart}
\setlength{\tabcolsep}{2.8pt}
\renewcommand{\arraystretch}{1.0}

\resizebox{\columnwidth}{!}{
\begin{tabular}{@{}lrrrrrr@{}}
\toprule
\textbf{Model}
& \textbf{\makecell{Lesion\\recogn.}}
& \textbf{\makecell{Binary\\condition}}
& \textbf{\makecell{Multi-disease}}
& \textbf{\makecell{DR grading}}
& \textbf{AMD}
& \textbf{Avg.} \\
\midrule

RETFound \cite{zhou2023foundation}
& 37.0 & 64.5 & 36.3 & 49.4 & 30.7 & 43.6 \\

DERETFound \cite{sun2025dataefficient}
& 40.4 & 62.5 & 34.2 & 49.2 & 33.5 & 44.0 \\

DINOv2-B \cite{oquab2024dinov2}
& 46.9 & 76.6 & 48.1 & 56.9 & 38.4 & 53.4 \\

DINOv2-L
& 47.7 & 75.5 & 45.7 & 59.1 & 37.9 & 53.2 \\

\midrule
Qwen2.5-VL-3B & 36.6 & 0.3 & 34.7 & 5.6 & 13.8 & 18.2 \\
InternVL2.5-8B & 46.6 & 1.2 & 41.3 & 30.9 & 32.5 & 30.5 \\
Qwen2.5-VL-7B
& 34.9 & 58.3 & 41.5 & 33.3 & 26.4 & 38.9 \\

% HealthGPT-M3-7B
% & 49.4 & 54.6 & 34.2 & 37.8 & 32.3 & 41.7 \\
%OphthaReasoner-Qwen-3B & 46.3 & 56.2 & 49.5 & 37.6 & 43.2 & 46.5 \\

%OphthaReasoner-Intern-2B & 45.7 & 61.0 & 45.5 & 40.1 & \textbf{45.3} & 47.5 \\

%Lingshu-7B & 53.7 & 52.3 & 40.8 & 53.8 & 39.0 & 47.9 \\

% HuatuoGPT-Vision-7B
% & 54.8 & 62.2 & \underline{50.4} & 41.4 & 36.9 & 49.1 \\

%FundusExpert-8B & \underline{60.6} & 50.0 & \textbf{61.5} & \textbf{65.2} & 22.5 & 51.9 \\

\rowcolor{blue!20}
Fundus-R1-3B
& \textbf{63.4} & \underline{79.8} & 34.4 & \underline{60.2} & 37.0 & \underline{55.0} \\
\rowcolor{blue!20}
Fundus-R1-7B
& 56.0 & \textbf{80.8} & 45.6 & 57.2 & \underline{41.1} & \textbf{56.1} \\

\bottomrule
\end{tabular}
}

\end{table}

\section{Conclusions}
\label{sec:conclusion}
We develop \model, a reasoning-enhanced fundus-reading MLLM trained using only public datasets. 
%Our central goal is to reduce the dependence of fundus-reading MLLMs on inaccessible in-house data and private clinical reports, while still enabling effective reasoning-oriented post-training under predominantly image-level supervision. To achieve this, we proposed a RAG-based reasoning-trace construction pipeline that combines image-specific visual findings with label- and modality-conditioned ophthalmic knowledge, and further incorporated an answer-dependent process reward into RLVR to improve the self-consistency of generated reasoning traces. 
Experiments on FunBench, Omni-Fundus, and GMAI-Fundus showed that \model consistently surpasses multiple strong baselines. These results indicate that reasoning supervision can be effectively induced from public data and can substantially improve fundus-reading performance, especially on knowledge-intensive tasks such as lesion analysis and disease diagnosis. \model takes an important step  towards democratizing medical MLLM research.

%We hope that this work will encourage more reproducible and accessible research on fundus-reading MLLMs.

%\medskip

\textbf{Limitations}.
Due to computational constraints, we use Qwen2.5-VL (3B/7B) as our base model. We ignore the influence of low-quality images, as they should ideally be flagged early, before fundus image reading commences.

%\yc{VF extraction could be further improved with specialized lesion segmentation models, while the model-agnostic design can generalize to post-training other generic MLLMs for fundus reading.}
%\yc{As our study focuses on downstream diagnostic after acquisition-side quality control, we include only fundus images that pass image-quality screening.}

{\small
\bibliographystyle{ieeenat_fullname}
\bibliography{11_references}
}

\ifarxiv \clearpage \appendix
\fi

\end{document}